# Transformer-based approach for Ethereum Price Prediction Using Crosscurrency correlation and Sentiment Analysis

Shubham Singh[1] , Mayur Bhat

1. New York University, New York

Abstract: The research delves into the capabilities of a transformer-based neural network for Ethereum cryptocurrency price forecasting. The experiment runs around the hypothesis that cryptocurrency prices are strongly correlated with other cryptocurrencies and the sentiments around the cryptocurrency. The model employs a transformer architecture for several setups from single-feature scenarios to complex configurations incorporating volume, sentiment, and correlated cryptocurrency prices. Despite a smaller dataset and less complex architecture, the transformer model surpasses ANN and MLP counterparts on some parameters. The conclusion presents a hypothesis on the illusion of causality in cryptocurrency price movements driven by sentiments.

1. Introduction

Cryptocurrencies have evolved into a significant investment asset class, witnessing exponential market capitalization growth from millions to billions (Pele et. al, 2023). Prominent digital currencies like Bitcoin and Ethereum have garnered substantial value, attracting widespread attention. This shift presents a distinctive opportunity to garner profound market insights and effectively forecast cryptocurrency prices.

The value of a coin is intricately tied to the sentiments and actions of its holders and investors (Koutmous et. al, 2023). Consequently, comprehending investor sentiment becomes crucial for accurately predicting price dynamics. To harness and analyze market sentiments, we propose a model that leverages data extracted from influential social media platforms such as Twitter and Reddit. By employing natural language processing (NLP) techniques and training deep learning models, our objective is to extract valuable insights from the market's collective sentiment.

The generated market insights will be combined with key features of the cryptocurrency market to predict future market values. To minimize errors and maximize prediction accuracy, the work considers the correlation between the target cryptocurrency and two highly correlated cryptocurrencies. This approach leverages the interdependencies among these coins to enhance the predictive capabilities of the model.

The main objective of this research paper is to present a comprehensive approach to Ethereum price prediction base on the assumption that Cryptocurrency prices are driven by sentiment analysis. By analyzing sentiments expressed on social media platforms, we can gain a deeper understanding of market dynamics and investor behavior.

## 2. Overview of related work

Cryptocurrencies have experienced significant price volatility over time, making their prediction challenging and even so valuable for trading. Numerous studies have explored the use of machine learning algorithms, deep learning models, and sentiment analysis to forecast cryptocurrency prices. However, some of the existing research suffers from several limitations and gaps in considering various factors and employing advanced techniques.

In 2017, works by Stanford researchers (Lamon et. al, 2017) employed machine learning algorithms to predict the prices of Bitcoin, Litecoin, and Ethereum based on news and social media sentiments. However, the research lacked exploration of advanced models like LSTM and GRU and had limited data consideration.

In 2017, another research (Radityo et. al, 2017) focused on using different variants of artificial neural networks (ANN) to predict Bitcoin prices. Although the study explored ANN methods, it did not consider sentiments or complex patterns using deep learning models.

Regression techniques, including Theil-Sen regression, Huber regression, LSTM, and GRU, were implemented for Bitcoin price prediction in 2018 (Phaladisailoed et. al, 2018). However, this research did not incorporate influencing factors like sentiments or hybrid models.

In 2019, a study utilized hidden Markov models and long short-term memory (LSTM) models for predicting cryptocurrency prices (Hashish et. al, 2019). However, the research did not consider market sentiment as a feature, which is important for prediction.

In 2020, a research paper introduced a big data platform for price prediction using sentiment analysis but did not include deep learning models (Mohapatra et. al, 2020). Another study utilized a hybrid LSTM-GRU model for price prediction but did not explore interdependencies among cryptocurrencies and sentiment as a feature (Kim et. al, 2021).

Various other studies conducted between 2020 and 2022 explored different models, including ARIMAX, CNN LSTM, hybrid LSTM-GRU, and ensemble models (Serafini et al., 2020)(Tanwar et al., 2021)(Politis et al., 2021).

However, several limitations were identified in the previous works such as the exclusion of market sentiments and interdependencies among cryptocurrencies.

## 3. Proposed work

The proposed model in this paper aims to address the limitations highlighted in previous works mainly a lack of sentiment analysis as a factor in the models. By incorporating sentiments from Twitter, Reddit, and news from CoinMarketCap, while considering factors such as price, transaction history, volume, and block size of the cryptocurrency, the model tries to improve upon the existing works.

Another addition to the model is using a robust deep learning architecture called transformers (Vaswani et. al, 2017). Transformers are known for their ability to retain context, uncover complex patterns, and generate outputs of similar complexity.

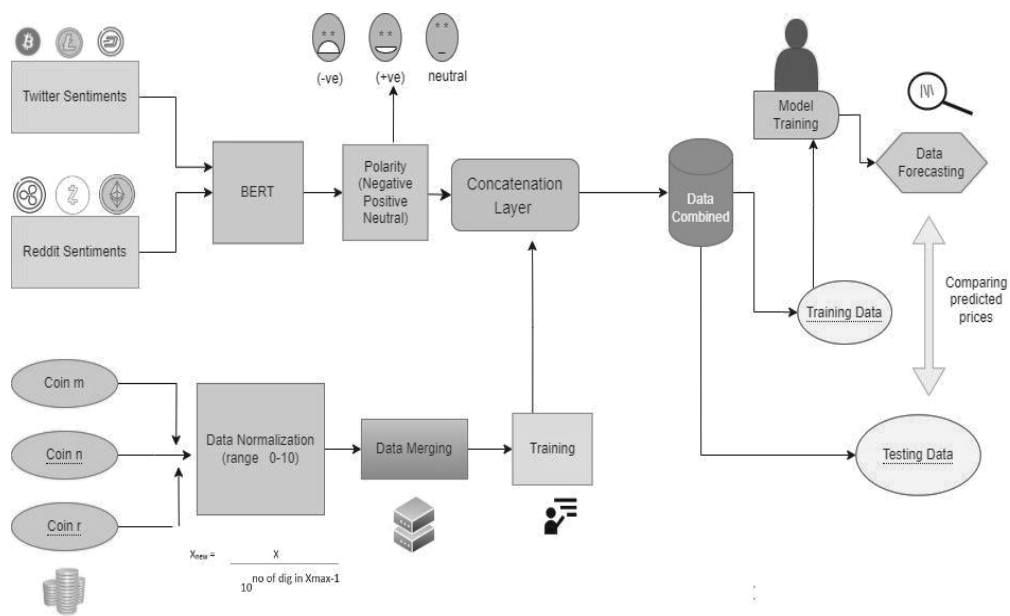

Figure 1: Flowchart of the proposed model

## 4. Data

The most important thing to note is the choice of temporal frame for the training data (~730 days) for Ethereum. The limited timeframe of the data is attributed to the availability of sentiment data for the time frame.

Coin prices and volume data are extracted from Yahoo Finance for different cryptocurrencies and normalized for each coin using the following formula:

$$normalized\ price = \frac{price}{10^{no.\ of\ digits}\ p\ max}$$

$$normalized\ volume = \frac{volume - minimum\ volume}{maximum\ volume - minimum\ volume}$$

The coins with the highest price movement correlations are selected for further analysis. This work uses Polkadot and Cardano interactions with Ethereum for price forecasting of Ethereum.

Sentiment data related to the selected cryptocurrencies are collected from social media platforms Twitter and Reddit. The collected data is then processed using a pre-trained transformer model called FinBert, specifically designed for sentimental analysis of financial data. FinBert (Araci, 2019) provides one hot encoded sentiment score from 0 to 1 for each encoding of positive, negative, and neutral.

To further normalize the sentiment scores, the mean of the positive, neutral, and negative scores is calculated. A sentiment score (score) is derived using the formula:

$$score = \frac{(positive + 0.5 * neutral)}{(positive + neutral + negative)}$$

This formula combines the counts of positive and neutral sentiments, considering half the weight of neutral sentiments, and divides it by the total sum of sentiments.

a. **Exploratory Data Analysis**

Correlation analysis of the data in Figure 2. highlights a strong correlation between pairs DOT-BTC (0.96), BNB-ETH (0.95), ETH-BTC (0.92), ETH-MATIC (0.91), BTC-MATIC (0.91). Suggesting a strong price movement in one when the other moves. For the experiment, ADA and DOT were taken as variables to predict ETH (Ethereum prices) due to reasons concerning the availability of sentiment-related data.

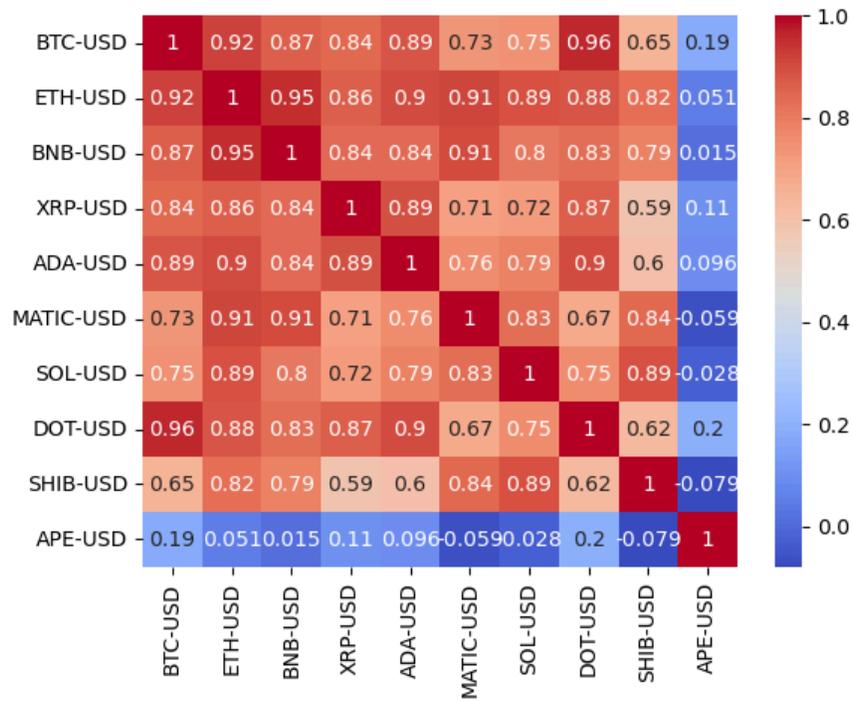

Figure 2: Heat map of correlation among various cryptocurrencies.

Figure 3. And Figure 4. Highlight a strong correlation between ETH, ADA, and DOT price and volume movements. Notably, there's a divergence in price movement after the FTX collapse.

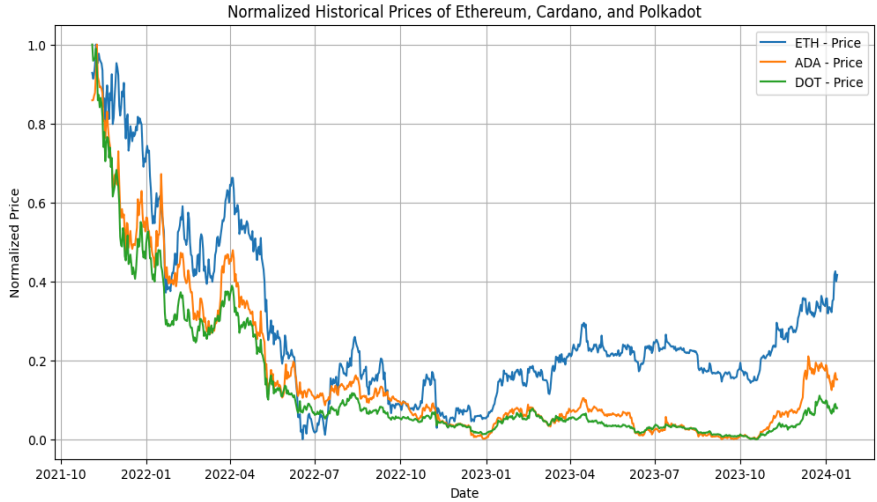

Figure 3: Normalized historical 800-day prices of ETH, ADA, DOT

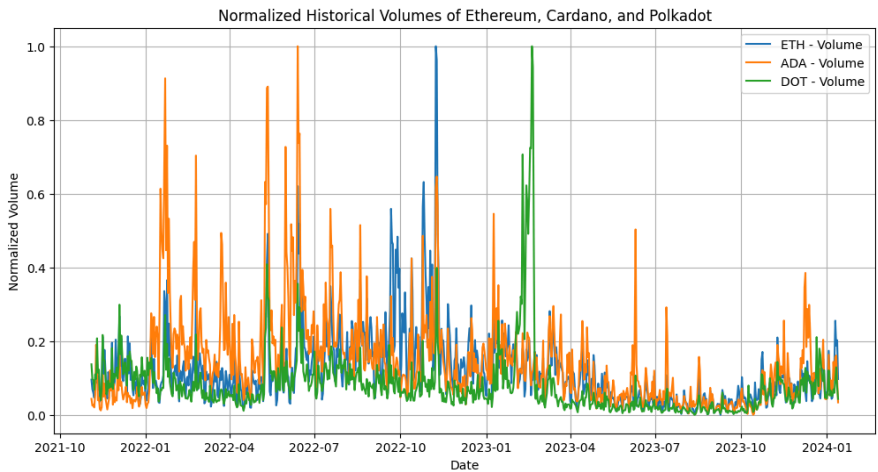

Figure 4: Normalized historical 800-day volumes of ETH, ADA, DOT

An exploratory analysis of the data suggests a significant correlation and similarities in the part to further continue the work.

## 5. Experiment

The model architecture is based on a transformer-based neural network consisting of transformer encoder blocks, each incorporating normalization, attention, and feed-forward mechanisms. The initial layers normalize input data and employ multi-head attention to capture intricate patterns, followed by convolutional operations for representation refinement. Stacking multiple transformer encoder blocks enhances the model's capacity to learn complex patterns, with global average pooling facilitating dimensionality reduction. However, nothing new has been employed in terms of model architecture or design. It is a standard transformer.

The experimental setups explore the model's performance under different feature configurations:

A single-feature configuration containing only Ethereum closing price data serves the purpose of establishing a baseline understanding of the model's performance when relying solely on temporal price data. By isolating Ethereum closing prices, the experiment aims to test the model's ability to capture patterns and trends without the complexity introduced by additional variables.

The experiment extends by including Ethereum price, volume, and sentiment data. The objective is to capture a more comprehensive array of dimensions, allowing the model to leverage features to improve decision-making. The proposed experimental setup in this paper is completed by adding prices of correlated coins such as ADA and DOT. Aiming to capture the correlation of the currencies as a feature to improve prediction capability.

The training data available in all cases is 579 days and the model is tested on 141 days of data. The small dataset is a factor that must be considered while assessing performance. All the models used use a mean squared error loss function to minimize loss while calculating gradients and use the Adam optimizer. The experiment includes 3 different data setups with a common model setup.

## 6. Result

Our transformer-based architectures utilizing Ethereum data give an MSE of 0.0051, 2.59 times better compared to LSTM scores observed just with Ethereum price (Kumar et. al, 2020). However, LSTM performs better in RMSE and MAPE scores than our model across all experiments. However, our model outperforms ANN and MLP in performance. Notably, the complexity and size of our dataset were significantly smaller compared to other works on similar topics. The result of this experiment concludes that Transformer displays some significance in predicting Ethereum prices. Below figure 5. show the predicted results compared to the actual value of the Ethereum price.

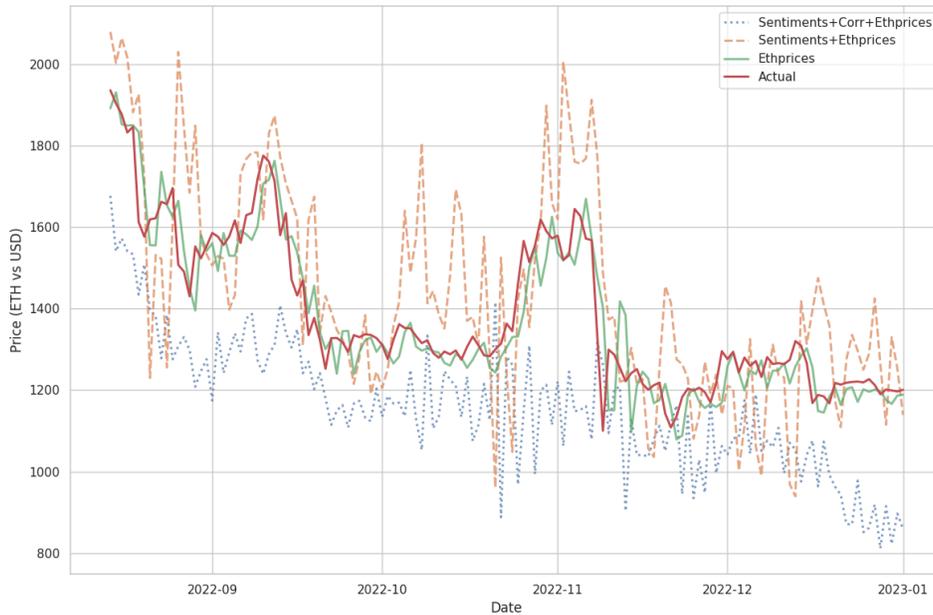

Figure 5: Ethereum prices predicted from experimental setup comparison to actual.

## Table 1. Model and Errors

| Model | RMSE | MSE | MAPE |
|---|---:|---:|---:|
| ANN (Kim et. al, 2021) | 0.068 | Not available | 0.048 |
| MLP (Kumar et. al, 2020) | 0.114 | 0.021 | 32.29 |
| LSTM (Kumar et. al, 2020) | **0.013** | 0.018 | **3.67** |
| Transformer + ETH data | 0.0716 | **0.0051** | 14.91 |
| Transformer + ETH data + Sentiments | 0.1892 | 0.0358 | 18.84 |
| Transformer + ETH data Cross-correlation data + Sentiments | 0.2608 | 0.068 | 18.14 |

## 7. Conclusion and Future Scope

The results display some notable results, which form the basis for motivation to work further on the topic. Despite the sharp correlation in price, volume, and sentiments with each other and with other cryptocurrencies, there's a low predictive power in the data, which lays the hypothesis of the illusion of causality in the claim that cryptocurrency price movements are driven by sentiments. However, the lack of large enough data to carry our experiments, especially on the sentiment side poses a challenge to verifying the hypothesis.

The focus going forward lies on collecting more data about sentiments from sources such as news articles and using the entire content of Reddit posts, and comments instead of just descriptions and content of articles as well. The work also calls for the use of existing models such as TST, Autoformer, and TimeGPT-1 and new time series transformers to capture data interactions and time series movements better. Another notable observation is that normalized price movements before the FTX collapse were more correlated with each other than now. This calls for an investigation of the effect of the event as a future work on the topic.

**Appendix:**

Code available: Ethereum-Price-Prediction